\pdfoutput=1

\documentclass[11pt]{article}

\usepackage{acl}

\usepackage{times}
\usepackage{latexsym}

\usepackage[T1]{fontenc}

\usepackage[utf8]{inputenc}

\usepackage{microtype}

\usepackage{inconsolata}
\usepackage{amsmath}
\usepackage{amsfonts}
\usepackage{graphicx}
\usepackage{subcaption}
\usepackage{enumitem}

\newcommand{\method}{FASST}

\title{\method: Fast LLM-based Simultaneous Speech Translation}

\author{
    Siqi Ouyang\textsuperscript{\dag}, Xi Xu\textsuperscript{\dag}, Chinmay Dandekar\textsuperscript{\ddag}, Lei Li\textsuperscript{\dag} \\
    \textsuperscript{\dag}Carnegie Mellon University
    \textsuperscript{\ddag}University of California, Santa Barbara \\
    \texttt{\{siqiouya, xixu, leili\}@cs.cmu.edu} \\
    \texttt{cdandekar@ucsb.edu}
}

\begin{document}
\maketitle

\begin{abstract}

Simultaneous speech translation (SST) takes streaming speech input and generates text
 translation on the fly.
Existing methods either have high latency due to recomputation of input representations, or fall behind of offline ST in translation quality.
In this paper, we propose \method, a fast large language model based method for streaming speech translation.
We propose blockwise-causal speech encoding and consistency mask, so that streaming speech input can be encoded incrementally without recomputation.
Furthermore, we develop a two-stage training strategy to optimize \method~for simultaneous inference.
We evaluate \method~and multiple strong prior models on MuST-C dataset.
Experiment results show that \method~achieves the best quality-latency trade-off.
It outperforms the previous best model by an average of 1.5 BLEU under the same latency for English to Spanish translation.



\end{abstract}
\section{Introduction}

End-to-end simultaneous speech translation (SST) translates incomplete speech input into text in a different language~\cite{ma-etal-2020-simulmt}, which is widely used in multilingual conferences, live streaming and etc. Compared to offline ST where speech input is complete, SST needs to decide whether to continue waiting or to generate more translation after receiving new speech input. A common approach in building performant SST streaming models involves pretraining for offline translation and optional finetuning for simultaneous translation~\cite{ agrawal-etal-2023-findings, communication2023seamless}. The quality-latency trade-off of simultaneous streaming models thus heavily depends on its offline performance. 

\begin{figure}[t]
    \centering
    \includegraphics[width=1\linewidth]{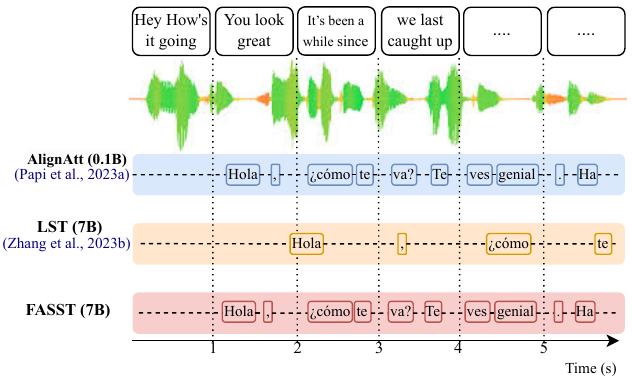}
    \caption{Simultaneous speech translation with AlignAtt-0.1B, LST-7B and our \method-7B. The LST-7B model generates translation with significantly higher latency than AlignAtt, while our \method-7B achieves comparable latency with it.
    }
    \label{fig:intro}
\end{figure}

Large language model (LLM) have recently demonstrated its potential to be a strong backbone of offline E2E ST~\cite{huang2023speech,zhang2023tuning}. However, LLM introduces larger computation overhead compared to regular-sized models when applied to SST. Figure \ref{fig:intro} shows that the computation latency of a LLM-based 7B model makes it inferior for real-time application.

The computation overhead of SST models comes from both encoding new speech input and decoding new translation. While the latter one has been heavily optimized for LLM~\cite{pope2022efficiently,kwon2023efficient,dao2024flashattention}, the former one has not been optimized for SST. As new speech input arrives, most SST models re-encode the entire speech and start autoregressive decoding afterwards, ignoring the incremental nature of streaming speech input. More importantly, the LLM decoder needs to recompute hidden states due to the updated speech features, significantly slowing down the computation.

In this work, we propose a \textbf{FA}st LLM-based \textbf{SST} (\method) method to avoid recomputation while maintaining its translation quality. We develop a blockwise-causal speech encoding technique that incrementally encodes new speech input and introduce incremental LLM decoding with consistency mask. We also design an 2-stage training strategy for \method: 1) aligning speech encoder outputs with LLM embeddings using word-aligned contrastive loss~\cite{ouyang-etal-2023-waco} and 2) finetuning for SST using wait-$k$-stride-$n$ policy~\cite{zeng-etal-2021-realtrans}. Experiments on MuST-C dataset~\cite{di-gangi-etal-2019-must} shows that our 7B model maintains competitive computation aware latency compared to 115M baselines while achieving consistent quality improvement of at least 1.5 BLEU score on English-Spanish direction.

Our contributions are:
\begin{itemize}[itemsep=1pt, leftmargin=10pt, parsep=0pt, topsep=1pt]
    \item We propose \method, one of the first efficient LLM-based methods for simultaneous speech translation.
    \item We verify \method~on MuST-C dataset and it outperforms strong prior methods by 1.5 BLEU at the same latency on English-Spanish direction.
    \item We further demonstrate that \method~can be generalized to other policies like hold-$n$ and policies spending more time on encoding benefit more from \method.
\end{itemize}

\section{Related Works}

\noindent\textbf{End-to-End SST}~ 
translates partial speech input into text in another language without generating intermediate transcription. 
A variety of speech segmentation techniques and policies have been proposed to optimize the quality-latency trade-off. \citet{ren-etal-2020-simulspeech,dong-etal-2022-learning,zeng-etal-2023-adatrans,zhang-etal-2023-training} learn to segment streaming speech input by word boundaries. \citet{zhang-feng-2023-end} further learns to segment speech at moments that are beneficial to the translation. On the policy side, \citet{ma-etal-2020-simulmt} adapts wait-$k$ and monotonic multihead attention (MMA) from simultaneous text translation to SST model. \citet{ma2023efficient} further improves the numerical stability of MMA. \citet{Papi2023} constructs source-target alignment with attention information to guide the simultaneous inference. \citet{zhang-feng-2022-information} decides whether to translate based on accumulated information of source speech. \citet{polak23_interspeech} conducts blockwise beam search when doing incremental decoding. The translation quality of SST models depend on not only their policies, but also their offline performance~\cite{agrawal-etal-2023-findings}. Recently LLM has been shown as a strong backbone of offline ST~\cite{zhang2023tuning,huang2023speech}, but its computation overhead prevents it from being used in SST scenarios. \method~is one of the first LLM-based SST models with a reasonable quality-latency trade-off.



\noindent\textbf{Efficient ST}~ To reduce the computation cost of ST models, \citet{wu2020streaming, ma2020streaming, raffel-chen-2023-implicit, raffel2023shiftable} use segments and explicit or implicit memory banks to calculate self-attention only within the segment. \citet{zhang-feng-2023-end, chen-etal-2021-direct, wu2021temporally} adopt unidirectional attention during speech encoding. These methods focus on encoder-side optimization and can be integrated with \method.

\noindent\textbf{Translation with LLM}~ While LLMs are capable of zero-shot machine translation \cite{brown2020language, openai2023gpt,touvron2023llama, llama2}, their performance can be further improved via in-context learning~\cite{vilar-etal-2023-prompting, 10.5555/3618408.3620130}, supervised and semi-supervised finetuning~\cite{rothe-etal-2020-leveraging, yang2023bigtranslate, zhang2023bayling,xu2023paradigm}. For simultaneous machine translation (SMT), \citet{guo2024sillm} propose a collaborative translation model with two LLM agents and \citet{koshkin2024transllama} design a finetuning strategy by adding a special "wait" token. \citet{raffel2024simultaneous} propose SimulMask to mask token connections under certain policy. SimulMask is a concurrent work with us and only works on text translation.


\begin{figure*}[htbp]
    \centering
    \includegraphics[width=1.0\textwidth]{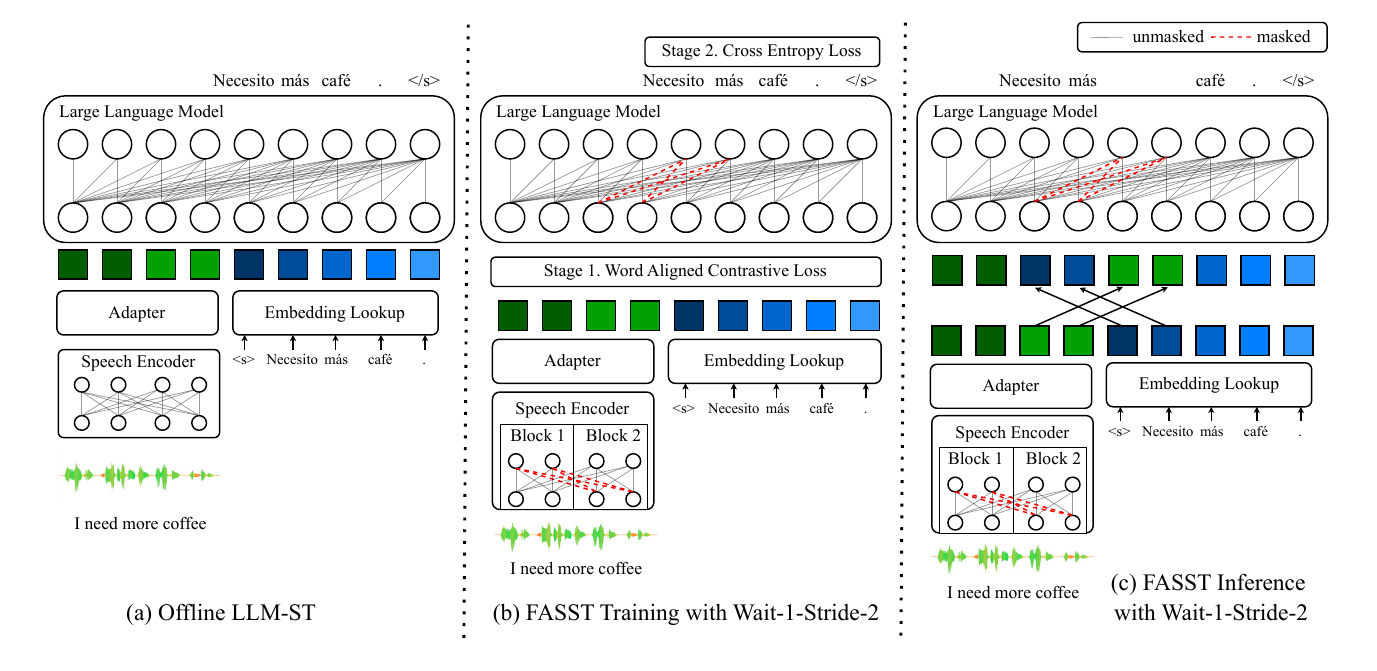}
    \caption{Overview of \method. (a) shows the offline translation of LLM-based ST model. (b) depicts the 2-stage training pipeline of \method. Stage 1 aligns adapter output with LLM embedding and stage 2 finetunes for simultaneous translation using wait-$k$-stride-$n$ policy. (c) illustrates the simultaneous inference procedure of \method~with incremental speech encoding and LLM decoding with consistency mask.}
    \label{fig:arch}
\end{figure*}

\section{The \method~Method}

In this section, we first review the problem formulation of simultaneous speech translation (SST) and then describe the architecture of our proposed model, \method, followed by its training and inference strategies.

\subsection{Problem Formulation}

Simultaneous speech translation (SST) needs to generate translations while receiving streaming speech input. Let $S=(s_1,s_2,\cdots,s_{|S|})$ be a speech waveform where $s_i$ are real numbers. The streaming speech input is cut into segments $S_1,S_2,\cdots$ and the SST model $P_{\theta}$ needs to emit partial translations $T_1,T_2,\cdots$ after receiving each of them,

\begin{equation}
    T_i \sim P_\theta (\cdot~|~S_{\leq i},T_{<i}).
\end{equation}
$T_i$ can be an empty string, indicating that the SST model needs more speech input to continue the translation. After receiving all inputs $S_1,S_2,\cdots,S_m$ and emitting all translations $T_1,T_2,\cdots,T_m$, we obtain the final translation $T=\bigoplus_{i=1}^m T_i$ by concatenating all partial ones.  

The objective of SST is to emit high-quality translation at a low latency. Quality is evaluated by comparing generated $T$ with the ground-truth $T^*$, while latency is evaluated based on the amount of lagging of each generated word. In this paper, we consider the computation-aware latency of SST models. 

\subsection{Model Architecture}

As shown in Figure \ref{fig:arch}, our model is composed of a speech encoder, an adapter and a LLM decoder.

\paragraph{Blockwise-Causal Speech Encoder (BCSE)} extracts contextualized acoustic features from the raw waveform incrementally. It consists of several casual convolutional layers as the audio feature extractor and a blockwise-causal Transformer Encoder as the contextual encoder. 

Our causal convolutional layers are built upon non-causal ones. Denote $H_{in}\in\mathbb{R}^{l \times d}$ as the input vectors to non-causal convolution $\text{Conv}(\cdot)$ with kernel size $w$. We add additional zero padding $\text{Pad} \in \mathbb{R}^{(w/2 - 1)\times d}$ to its left so that each output vector only depends on input vectors to its left, and remove the last $w/2 - 1$ states to keep its output length the same as before,
\begin{align}
    H_{out} = \text{Conv}\left( \text{Pad}\oplus H_{in} \right)_{:-w/2 + 1}.
\end{align}

Besides, we apply blockwise-causal masking to Transformer Encoder. Define attention mask $M$ of speech encoder as follows
\begin{align}
    M_{j_Q,j_K} = \begin{cases}
        0 & \left\lfloor\frac{j_Q}{b}\right\rfloor \geq \left\lfloor\frac{j_K}{b}\right\rfloor \\
        -\infty & \text{otherwise}
    \end{cases}
\end{align}
where $b$ is the block size, i.e., the number of hidden states of the speech encoder corresponding to one segment, and $j_Q,j_K$ are row indices of query matrix $Q$ and key matrix $K$.
The attention output of speech encoder during training can then be written as
\begin{align}
    O = \text{Softmax}\left( \frac{Q K^T}{\sqrt{d}} + M \right)V,
\end{align}
where $V$ is the value matrix.

\paragraph{Adapter} receives speech encoder outputs and converts them to the LLM embedding space. It consists of two causal convolutional layers to reduce the length of speech encoder outputs by four and one linear layer to project features into the LLM embedding space. We call the adapter outputs as speech embeddings,
\begin{align}
    E^s_{\leq i} = \text{Adapter}(\text{BCSE}(S_{\leq i})).
\end{align}

\begin{figure}
    \centering
    \includegraphics[width=0.9\linewidth]{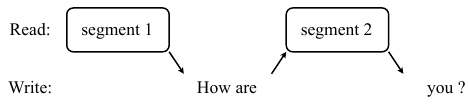}
    \caption{Example of wait-$1$-stride-$2$. It waits for $1$ segment at the beginning and then alternate between generate $2$ words (including punctuations) and reading new segment.}
    \label{fig:wait-k-stride-n}
\end{figure}

\paragraph{LLM} receives speech embeddings and embeddings of previously generated tokens to decode autoregressively according to a wait-$k$-stride-$n$ policy $\pi$.
\begin{align}
    T_i\sim \text{LLM}(\cdot~|~E^s_{\leq i}, T_{<i}, \pi).
\end{align}
Wait-$k$-stride-$n$ policy waits for $k$ speech segments at the beginning and then alternate between generating $n$ words and reading new segment. Figure \ref{fig:wait-k-stride-n} shows an example of wait-$1$-stride-$2$.

\subsection{Training}

As shown in Figure \ref{fig:arch} (b), we employ a 2-stage approach to train our model. 

\paragraph{Stage 1. Speech-text alignment.} We align the speech embedding with LLM input embedding using word-aligned contrastive (WACO) loss. Both transcription embeddings $E^{t}$ and speech embeddings $E^s$ are grouped into word embeddings $W^t$ and $W^s$ by word boundaries. Word boundaries of speech are obtained through Montreal Forced Aligner~\footnote{\url{https://github.com/MontrealCorpusTools/Montreal-Forced-Aligner}}. We treat speech and transcription embeddings of the same word as positive pair and others as negative pairs and train the speech encoder and the adapter with contrastive loss,
\begin{align}
    \mathcal{L}_\text{CTR} = -\mathbb{E}_{i}\left[\log \frac{\exp(sim(W^s_i,W^t_i)/\tau)}{\sum_{j}\exp(sim(W^s_i,W^t_j)/\tau)}\right]
\end{align}
where $\tau$ is the temperature and $sim()$ is the cosine similarity function. LLM parameters are frozen during stage 1.

\paragraph{Stage 2. Finetuning for simultaneous translation.} We finetune the entire model for simultaneous speech translation using wait-$k$-stride-$n$ policy. Speech input is encoded into speech embeddings $E^s$. Then we concatenate $E^s$ with embeddings of reference translation and feed them to LLM. 
Position indices of both speech embeddings and translation embeddings start with the same index and ascend separately, so that text generation during inference does not affect the positional embeddings of speech embeddings.

Then we randomly select $k\in K$ and mask out attentions from translation words with indices from $in$ to $(i+1)n-1$ to speech segments $S_{>i+k}$ for each $i$, since these words are generated before speech segments $S_{>i+k}$ arrive during inference. Finally, we apply the cross-entropy loss to train the entire model,
\begin{align}
    \mathcal{L}_\text{CE} = -\mathbb{E}_i\left[ \text{LLM}(T_i^*~|~E^s,T_{<i}^*,\text{mask}) \right],
\end{align}
where $T^*$ is the reference translation. 

\subsection{Efficient Simultaneous Inference}

Figure \ref{fig:arch} (c) illustrates how we conduct efficient simultaneous inference. 
\method~ waits for $k$ segments at the beginning and then start generating. 
Suppose now we have received $S_1,S_2,\cdots,S_i$ where $i\geq k$.

\paragraph{Incremental Speech Encoding} 
The blockwise-causal mask of speech encoder allows us to use KV cache of previous speech segments to avoid recomputation. Let $H^s = \left(h^s_1, \cdots, h^s_{l_i}\right)$ be input vectors of the attention. We group them into blocks $B_j = \left(h^s_{(j - 1)b + 1}, \cdots, h^s_{jb}\right)$ where $1\leq j\leq i$ and $i\cdot b = l_i$. The query, key and value matrices can be written as follows 
\begin{align}
    Q = H^s M_Q &= (B_1M_Q, \cdots, B_iM_Q) \\
    K = H^s M_K &= (B_1M_K, \cdots, B_iM_K) \\
    V = H^s M_V &= (B_1M_V, \cdots, B_iM_V)
\end{align}
Here the keys and values of previous segments $(B_1M_K,\cdots,B_{i-1}M_K)$ and $(B_1M_V,\cdots,B_{i-1}M_V)$ are stored in the KV cache.
Now we only need the KV cache and the query $B_iM_Q$, key $B_iM_K$ and value $B_iM_V$ of the latest segment to compute its attention output,
\begin{align}
    O_i^s = \text{Softmax}\left( \frac{B_iM_Q K^T}{\sqrt{d}} \right)V.
\end{align}
This results in same output as running attention with full query, key and value matrices and a blockwise-causal mask. In this way, we reduce the time complexity of attention from $O(l_i d^2 + l_i^2d)$ to $O(b d^2 + l_i bd)$. Here $b$ is a constant while $l_i$ increases with the longer speech input.

\paragraph{Adapting} We store the speech encoder outputs of previous segments and concatenate them with encoder outputs of segment $i$. Then we pass them to the causal convolutional layers and the linear layer to obtain the speech embeddings $E^s_{\leq i}$. 

\paragraph{LLM Decoding with Consistency Mask} 
We partition speech embeddings $E^s_{\leq i}$ into $E^s_1,\cdots,E^s_i$ by speech segment.
Following the wait-$k$-stride-$n$, the inputs to LLM are organized in the follow way
\begin{align}
    I = E^s_1 \oplus \cdots \oplus E^s_k \oplus \text{Emb}(T_k) \oplus E^s_{k+1} \oplus \nonumber\\
    \text{Emb}(T_{k+1}) \oplus \cdots \oplus E^s_i,
\end{align}
where $T_{1:k-1}$ are empty strings and $T_{j}$ consists of $n$ words for each $k\leq j < i$.
Now we need to reuse KV cache of previous $i-1$ speech segments and partial translations to compute LLM hidden states of $i_{th}$ segment. Since speech embeddings are always ahead of text embeddings during training, we design a consistency mask to ensure speech segments can only attend to speech segments before them.

Let $\delta(z)$ be indicator function that equals to $1$ if $z_{th}$ position of input $I$ belongs to text and $0$ otherwise. Define consistency mask $M^c$ as follows,
\begin{align}
    M^c_{z_Q,z_K} = 
    \begin{cases}
         0 & z_Q\geq z_K\text{ and }\delta(z_Q)\geq \delta(z_K) \\
         -\infty & \text{otherwise} \\
    \end{cases}
\end{align}
Let $Q_i, K_i, V_i\in\mathbb{R}^{t_i\times d}$ be query, key and value matrices of segment $i$ and $K_{<i},V_{<i}$ be cached key and value matrices. 
We first concatenate $K_i$ and $V_i$ with cache to obtain $K_{\leq i}$ and $V_{\leq i}$.
The attention output of segment $i$ can then be computed as follows
\begin{align}
    O_i^t = \text{Softmax}\left( \frac{Q_i K_{\leq i}^T}{\sqrt{d}} + M^c_{-t_i:,:} \right)V_{\leq i}.
\end{align}

After computing hidden states for speech segment $S_i$, the LLM decodes $n$ words autoregressively following the policy.
\section{Experiment}

\begin{figure}
    \centering
    \includegraphics[width=0.85\linewidth]{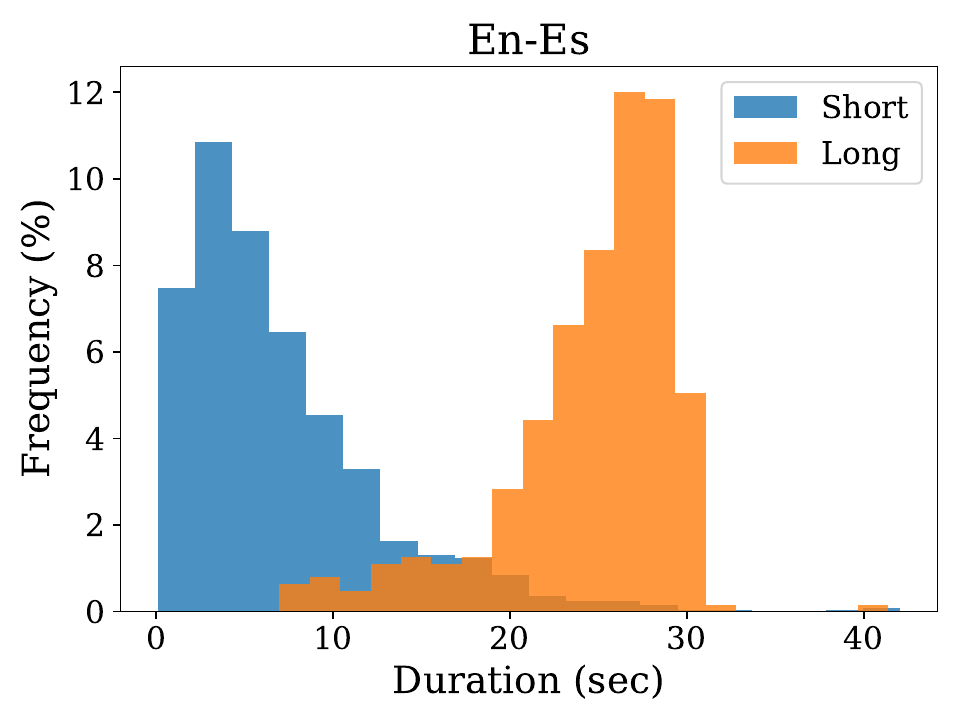}
    \caption{Duration distribution of MuST-C-Short and MuST-C-Long. The average duration of MuST-C-Short is around 5 seconds while that of MuST-C-Long is around 25 seconds.}
    \label{fig:data_dist}
\end{figure}

\subsection{Dataset}

We conduct experiments on two language directions of MuST-C v1.0 dataset \cite{di-gangi-etal-2019-must}: English$\to$Spanish (En-Es) and English$\to$German (En-De). Each language direction contains around 400 hours of audio recordings. The average duration of utterances is less than 10 seconds. To simulate long speech scenarios, we concatenate adjacent utterances in the same TED talk so that each resulting utterance is around 30 seconds. We call the induced dataset as MuST-C-Long\footnote{The manifest of MuST-C-Long will be released together with the code.} and the original one as MuST-C-Short. The duration distribution of both datasets are shown in Figure \ref{fig:data_dist}.

\begin{figure*}
    \centering
    \includegraphics[width=0.8\linewidth]{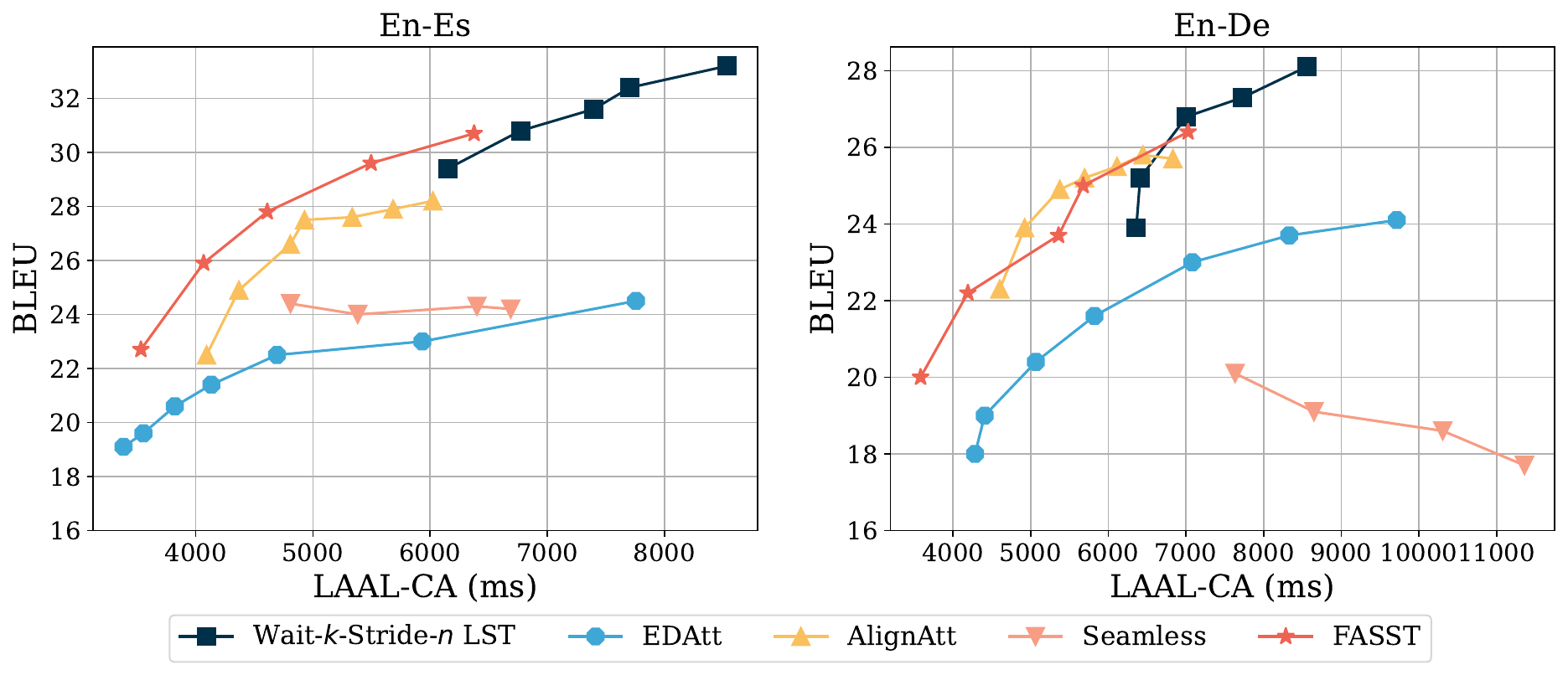}
    \caption{Quality-latency trade-off of \method~and baselines on English-Spanish and English-German direction. Quality is reflected by BLEU and latency is reflected by computation-aware length-adaptive average lagging (LAAL-CA). Given long speech input and large batch size, our model achieves overall the best quality-latency trade-off.}
    \label{fig:main}
\end{figure*}

\subsection{Model Configurations}

\paragraph{Architecture} We intialize our speech encoder with wav2vec 2.0 large model\footnote{\url{https://dl.fbaipublicfiles.com/fairseq/wav2vec/wav2vec_vox_960h_pl.pt}} ~\cite{baevski2020wav2vec} and our LLM with Llama2 7b Base\footnote{\url{https://huggingface.co/meta-llama/Llama-2-7b}}~\cite{touvron2023llama}. Wav2vec 2.0 large consists of a 7-layer convolutional feature extractor and a 24-layer Transformer encoder with 1024 hidden units. The block size of speech encoder is set to 50, i.e., around 1 second each block. The adapter connecting wav2vec 2.0 and Llama2 consists of two 1-D convolutional layers with kernel size 3, stride 2 and hidden size 1024 and a linear layer to project hidden size from 1024 to 4096 to match that of LLM embedding. Llama2 7b Base adopts a 32-layer Transformer decoder with hidden size 4096. It uses a vocabulary of size 32000 and rotary positional embedding \cite{su2023roformer}.

\paragraph{Training} We train our model with mixed MuST-C-Short and MuST-C-Long data. The input speech is raw 16-bit 16kHz mono-channel waveform. We filter out speech that is shorter than 320ms during training. The batch size of stage 1 is 16.7 minutes and that of stage 2 is 14 minutes. We use AdamW optimizer with cosine learning rate decay. The warmup steps of stage 1 is 25k and that of stage 2 is 500 steps. The maximum learning rate of stage 1 is 1e-4 and that of stage 2 is 2e-5. Gradients are clipped by the norm of 10. We train stage 1 for 500k steps and stage 2 for 1 epoch. We choose the checkpoint with the lowest dev loss. All our models are trained on 4 Nvidia A6000 GPUs with fp16. The temperature of WACO loss is 0.2 and we set $K=\{1,2,3,4,5,100\},n=3$ for stage 2 training.

\paragraph{Inference} We set speech segment size to 1 second to match the block size. We wait for $1\leq k\leq 5$ segments at first. Then as each segment arrives, the speech encoder encodes its speech embedding incrementally and pass it to LLM, where LLM computes hidden states without recomputation and generates $n=3$ words with greedy decoding as the partial translation.

\subsection{Evaluation}

We use SimulEval \cite{ma-etal-2020-simuleval} to evaluate our models and baselines. All models are evaluated on MuST-C-Long tst-COMMON with batch size of 8 during inference to simulate heavy workload. Since SimulEval does not support batching multiple instances, we duplicate each instance by 8 during model forwarding. We report SacreBLEU \cite{post-2018-call} for translation quality and computation-aware length-adaptive average lagging (LAAL-CA) \cite{papi-etal-2022-generation} for latency. All models are evaluated using a single A6000 GPU.

\subsection{Baselines}

\paragraph{Wait-$k$-Stride-$n$ LST} waits $k$ fixed-length speech segments and translates $n$ words every time \cite{ma-etal-2020-simulmt,zeng-etal-2021-realtrans}. We run wait-$k$-stride-$n$ policy on a strong offline LLM-based model LST \cite{zhang2023tuning} trained on the same mixed data as \method. LST has the Encoder-Adapter-LLM architecture similar to \method~but employs bidirectional speech encoder and requires recomputation every time a new speech segment arrives. We set $k\in\{1,2,3,4,5\}$, $n=3$ and segment length 1 second to match the setting of \method.
 
\paragraph{EDAtt} is an attention-based adaptive policy~\cite{papi-etal-2023-attention}. It leverages the encoder-decoder attention of an offline ST model to decide when to emit partial translations. The intuition is that if the attention is focused on early audio frames, the current translation can be emitted since sufficient information has been received. We use the model checkpoint and settings provided by the authors.

\paragraph{AlignAtt} is the current state-of-the-art (SOTA) method that extends EDAtt by explicitly generating audio-translation alignments from encoder-decoder attention \cite{Papi2023}. While EDAtt emits based on attention scores directly, AlignAtt decides based on whether a predicted token aligns with the latest audio frames, providing a more interpretable latency control. We also use the model checkpoint and optimal settings provided by the authors of AlignAtt.

\paragraph{Seamless} is a multilingual streaming speech translation system with efficient monotonic multihead attention mechanism~\cite{ma2023efficient} to generate low latency translation~\cite{communication2023seamless}. It computes target to source alignment using cross attention and writes translation if the alignment probability is larger than a threshold. We vary the threshold in $[0.2, 0.4, 0.6, 0.8]$ to evaluate its quality-latency trade-off.


\subsection{Main Results}

Main results are shown in Figure \ref{fig:main}. Our model achieves the best quality-latency trade-off for En-Es direction. Although wait-$k$-stride-$n$ LST has a 2 BLEU score advantage at the latency of 8 seconds, its bidirectional encoding and inefficient use of KV cache prohibit it reaching latency smaller than 6 seconds. Comparing to EDAtt and AlignAtt which do not use LLM and has much less parameters (115M) than our model (7B), our model has similar computation aware latency while achieving a 1.5 BLEU score improvement. For En-De direction, \method~achieves competitive results to AlignAtt, with slightly better quality when latency is smaller than 4 seconds or larger than 6 seconds.

\begin{figure}[t]
    \centering
    \includegraphics[width=0.85\linewidth]{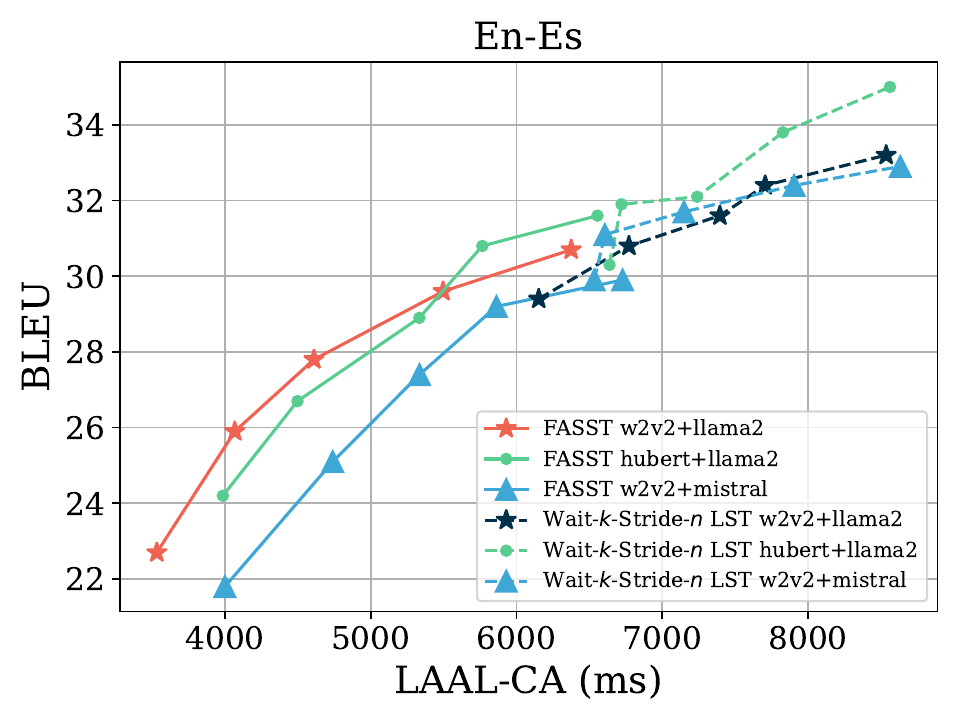}
    \caption{Ablation on the choice of pretrained speech encoder and LLM. We replace wav2vec 2.0 large with HuBERT large and Llama2 7B with Mistral 7B v0.3 base. \method~consistently has lower latency than Wait-$k$-Stride-$n$ LST while maintaining an acceptable translation quality.}
    \label{fig:ablation-s-enc-llm}
\end{figure}

\begin{figure}[t]
    \centering
    \includegraphics[width=0.9\linewidth]{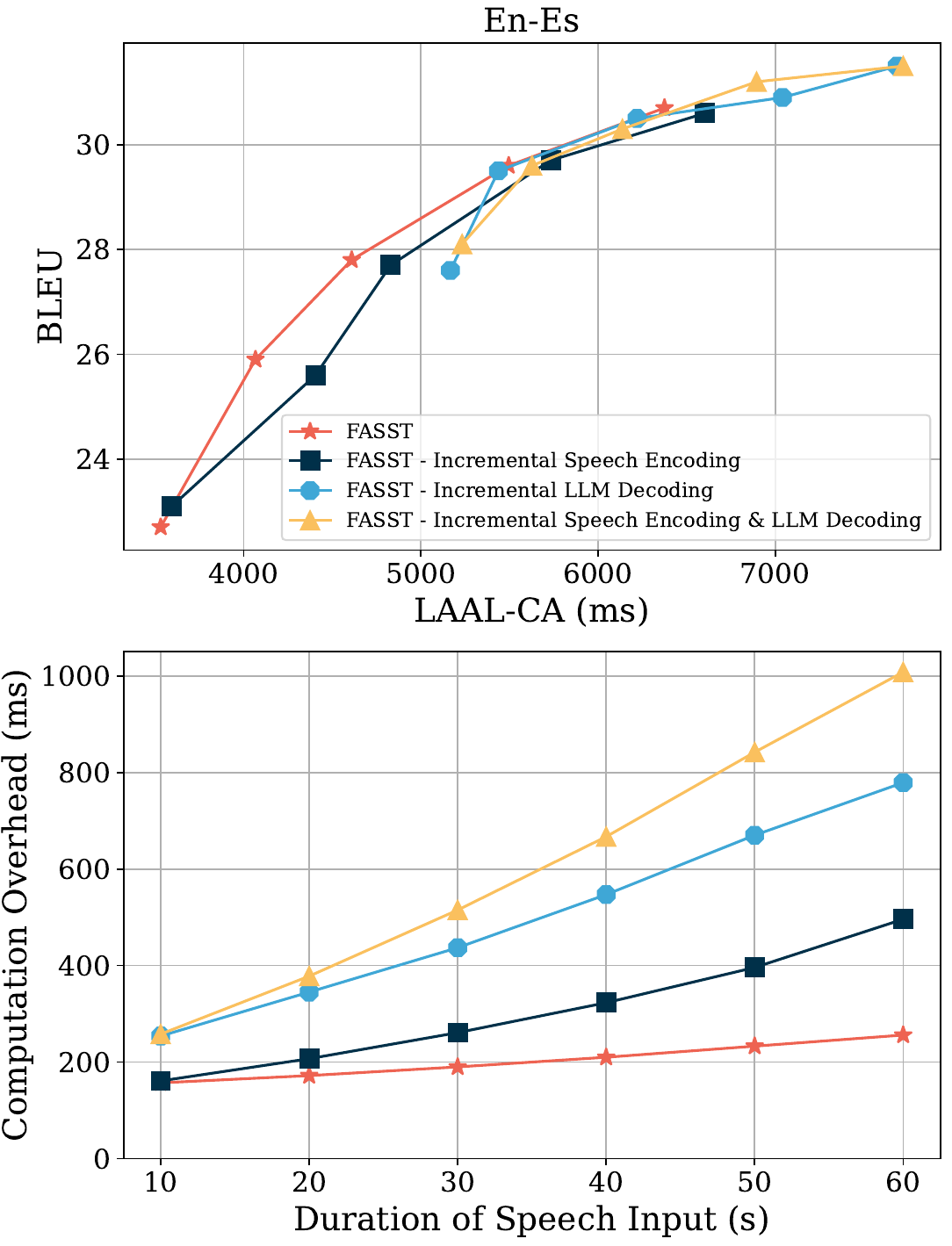}
    \caption{Ablation on incremental speech encoding and LLM decoding. Removing both incremental encoding and decoding can result in 4x larger latency for 60 seconds speech input.}
    \label{fig:ablation-incremental}
\end{figure}


    
\begin{figure}[t]
    \centering    
    \includegraphics[width=0.85\linewidth]{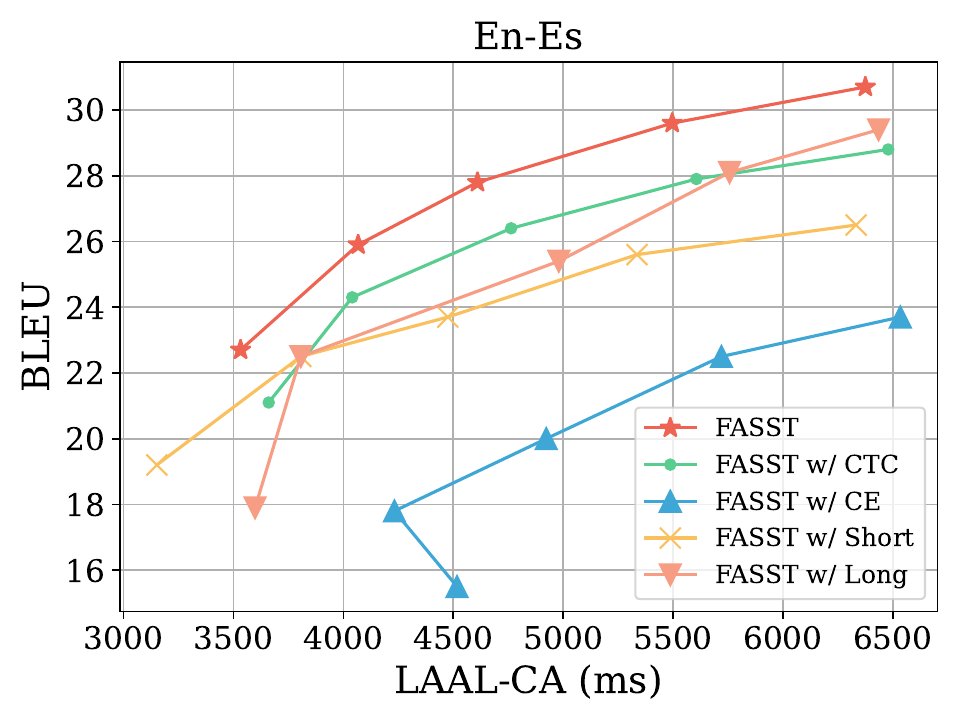}
    \caption{Ablation on the training strategy of \method. We train stage 1 WACO loss with CTC and cross entropy (CE) loss and also change the training data to only MuST-C-Short and only MuST-C-Long. WACO and mixed-data training achieves the best performance.}
    \label{fig:ablation}
\end{figure}

\subsection{Ablation Studies}

We conduct ablation studies to examine the impact of each component in our model.

\paragraph{Speech Encoder and LLM} We replace wav2vec 2.0 large with HuBERT large~\cite{hubert} and Llama2 7B base with Mistral 7B v0.3 base~\cite{jiang2023mistral} to examine whether \method~is sensitive to the choice of pretrained speech encoder and LLM. We also train Wait-$k$-Stride-$n$ LST baseline with these configurations as a comparison. Results are shown in Figure \ref{fig:ablation-s-enc-llm}. For all configurations, \method~has lower latency than the baseline. \method~with HuBERT results in the best quality when the latency is around 5.5$\sim$6.5 seconds and \method~with wav2vec 2.0 becomes the best when the latency is smaller than 5.5 seconds.

\paragraph{Incremental Encoding and Decoding} We ablate the incremental speech encoding and the incremental LLM decoding to examine their impact. For encoding, we use the same architecture but recompute the entire speech encoder at each step. For decoding, we recompute the entire LLM hidden states given updated speech input and then incrementally decode the translation as each speech segment arrives. This also provides translation tokens with more context since they can attend to speech embeddings appear after them. Results are shown in Figure \ref{fig:ablation-incremental}. Incremental encoding of speech encoder reduces the computational latency consistently by at least 200ms compared to recomputing encoder. Recomputing LLM does improve translation quality ($1\sim 5$ BLEU), but also introduces significant computation overhead ($\sim 1.5$ second), making it inferior for real-time application. 

We also plot the computation cost of each read/write step for each variant in Figure \ref{fig:ablation-incremental} with wait-$2$-stride-$3$ policy. \method~scales the best with the speech input length and reduces the overhead by at most 4x compared to the one without incremental encoding and decoding. Removing incremental decoding results in larger additional computation overhead compared to removing incremental encoding since LLM has far more parameters than the speech encoder.



\paragraph{Speech-Text Alignment Training} We replace WACO loss with CTC loss \cite{graves2006connectionist} and cross entropy loss in stage 1 to examine its impact on model performance. CTC loss aligns speech and text embeddings by optimizing all possible alignment paths. For cross entropy loss, we pass the speech embeddings to LLM and optimize the cross entropy loss with LLM parameters frozen. As shown in Figure \ref{fig:ablation}, WACO loss consistently outperforms CTC and cross entropy by at least 1 BLEU score at the same latency.

\paragraph{Mixing MuST-C-Long and Short} We train our model separately with only MuST-C-Short and only MuST-C-Long for the same number of epochs. As shown in Figure \ref{fig:ablation}, the model trained with both long and short data outperforms the one trained with short data by up to 4 BLEU points. Though we are using LLM, the length extrapolation is still unsatisfactory. Training with long data improves the quality compared to short data at high latency, but still outperformed by mixed-data training. 

\begin{figure}[t]
    \centering
    \includegraphics[width=0.85\linewidth]{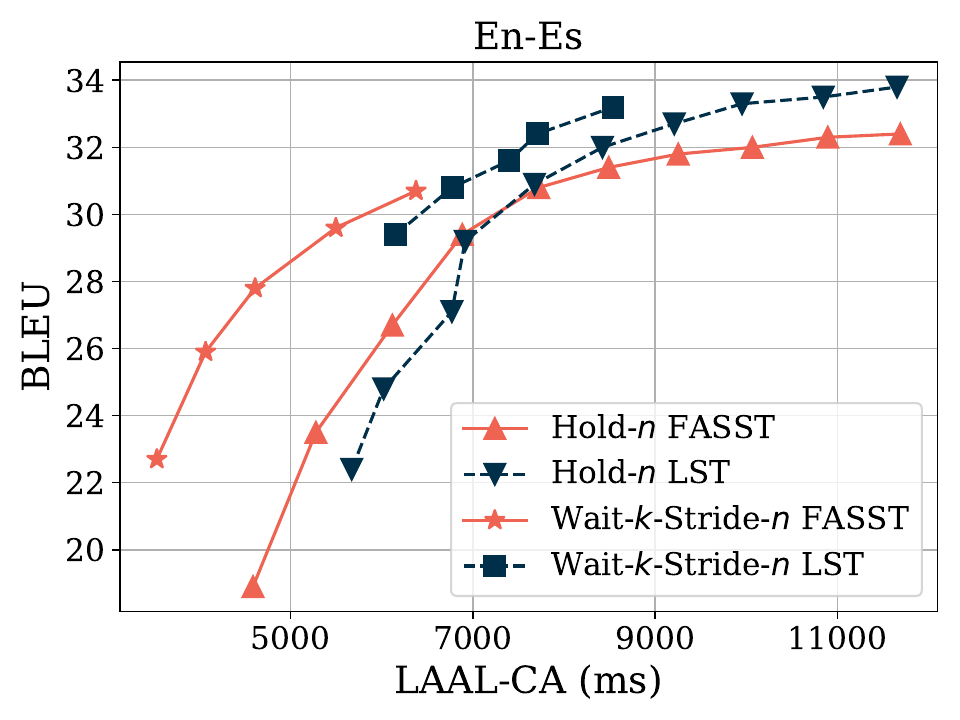}
    \caption{Quality-latency trade-off of \method~when applied to hold-$n$ policy. We observe less improvement with respect to LST than wait-$k$-stride-$n$ policy.}
    \label{fig:policy}
\end{figure}

\subsection{Generalizability to Other Policy}

We have demonstrated that our method works with wait-$k$-stride-$n$ policy.
However, plenty of policies other than wait-$k$ and its variants have been developed to conduct simultaneous translation. 
Here we apply our method to hold-$n$ policy~\cite{liu20s_interspeech} to exemplify how our method works on a different policy and in the meanwhile explain the factors that influence the effectiveness of our method.

Hold-$n$ policy selects the best hypothesis after each speech segment arrives, discard the last $n$ tokens from it and outputs the rest as the partial translation. Since hold-$n$ is a test-time policy for offline ST model, we train an offline version of our model by replacing stage 2 finetuning with standard offline finetuning using cross entropy loss but keep blockwise-causal encoding. During inference, we still conduct the same incremental encoding and decoding.  

As shown in Figure \ref{fig:policy}, our method has less advantage when applied to hold-$n$ comparing to wait-$k$-stride-$3$. The major difference between two policies in terms of computation is that hold-$n$ policy spends more time on autoregressive decoding since it decodes more tokens each time. On average hold-$n$ policy generates more than 6 words each step while wait-$k$-stride-$3$ generates at most 3 words. \method~accelerates the encoding of existing features, but for policies like hold-$n$ that involve heavy autoregressive decoding the advantage of our method gets marginalized.
\section{Conclusion}
In this work, we introduce \method, a fast LLM-based simultaneous speech translation model. \method~consists of blockwise-causal speech encoding, incremental LLM decoding with consistency mask, and a novel 2-stage training strategy. Experiments on MuST-C dataset show that \method~significantly reduce computation overhead while maintaining its translation quality. Our generalization study shows that policies that spend more time on encoding than decoding benefit more from \method.


\section*{Limitations}

\begin{itemize}[itemsep=1pt, leftmargin=10pt, parsep=0pt, topsep=1pt]
    \item There might be data leakage since LLM is trained on vast amount of text data, so we cannot guarantee LLM does not see the test translation data during pretraining.
    \item \method~is only tested on two language directions instead of all 8 language directions of MuST-C, so its generalizability to other language directions is unknown.
    \item There is still a quality gap between blockwise-causal speech encoding and bidirectional speech encoding. It is unclear how to further close the gap.
    \item We only explore one LLM-ST architecture in the paper and we cannot guarantee that \method~or its idea works on other architectures. 
\end{itemize}

\bibliography{anthology,custom}

\appendix

\end{document}